\documentclass[conference]{IEEEtran}
\IEEEoverridecommandlockouts
\usepackage{cite}
\usepackage{epsfig,rotating,setspace,latexsym,amsmath,epsf,amssymb,bm}
\usepackage{amsthm}
\usepackage{bbm}
\usepackage{color, hyperref}
\usepackage{multicol}
\usepackage{amsmath,amssymb,amsfonts}
\usepackage{algorithmic}
\usepackage{graphicx}
\usepackage{textcomp}
\usepackage{xcolor}
\usepackage{algorithmic}
\usepackage{algorithm}
\usepackage{caption}
\newtheorem{definition}{Definition}
\newtheorem{theorem}{Theorem}

\newtheorem{remark}{Remark}

\newtheorem{proposition}[theorem]{Proposition}

\def\BibTeX{{\rm B\kern-.05em{\sc i\kern-.025em b}\kern-.08em
    T\kern-.1667em\lower.7ex\hbox{E}\kern-.125emX}}
\usepackage[compact]{titlesec}         
\titlespacing{\section}{8pt}{8pt}{8pt} 
\AtBeginDocument{
  \setlength\abovedisplayskip{2pt}
  \setlength\belowdisplayskip{2pt}}
\begin{document}

\title{Learning Fair Robustness via Domain Mixup\\
\thanks{This work was supported by NSF grants CCF 2100013, CNS 2209951, CCF 1651492, CNS 2317192, CNS 1822071, by the U.S. Department of Energy, Office of Science, Office of Advanced Scientific Computing under Award Number DE-SC-ERKJ422, and by NIH Award 1R01CA261457-01A1.}
}

\author{\IEEEauthorblockA{Meiyu Zhong ~~ Ravi Tandon}
\IEEEauthorblockA{Department of Electrical and Computer Engineering \\
University of Arizona, Tucson, USA \\
E-mail: \textit{\{meiyuzhong, tandonr\}}@arizona.edu
}}

\maketitle

\begin{abstract}
Adversarial training is one of the predominant techniques for training classifiers that are robust to adversarial attacks. Recent work, however has found that adversarial training, which makes the overall classifier robust, it does not necessarily provide equal amount of robustness for all classes. In this paper, we propose the use of \textit{mixup} for the problem of learning \textit{fair} robust classifiers, which can provide similar robustness across all classes. Specifically, the idea is to mix inputs from the same classes and perform adversarial training on mixed up inputs. We present a theoretical analysis of this idea for the case of linear classifiers and show that mixup combined with adversarial training can provably reduce the class-wise robustness disparity. This method not only contributes to reducing the disparity in class-wise adversarial risk, but also  the class-wise natural risk. Complementing our theoretical analysis, we also provide experimental results on both synthetic data and the real world dataset (CIFAR-10), which shows improvement in class wise disparities for both natural and adversarial risks.

\end{abstract}

\section{Introduction}
 Despite their advanced capabilities, modern neural networks are known to be vulnerable to adversarial attacks. These attacks involve subtle, almost imperceptible modifications to the input data—known as adversarial examples—that can lead the networks to make incorrect decisions \cite{goodfellow2014explaining, szegedy2013intriguing,morgulis2019fooling}. Therefore, to defend against adversarial attacks, numerous researchers have proposed  algorithms for learning classifiers that are robust to adversarial attacks. 
Madry et al. \cite{madry2017towards} introduced adversarial training (AT) to enhance the robustness of models against adversarial examples. Informally, the goal of AT is to minimize the robust risk (or a surrogate of robust risk) as opposed to the conventional natural risk. Several variations of AT have been developed. For instance, research works \cite{pang2020boosting,zhang2019theoretically} introduce different regularization terms, while other works \cite{carmon2019unlabeled,schmidt2018adversarially} incorporate unlabeled data into the training set. Additionally, Wong et al. \cite{wong2020fast} present a method to expedite the training process. Extensive research \cite{dobriban2020provable,javanmard2020precise,zhong2024splitz,zhang2024filtered} have explored the balance between robustness and accuracy. Notably, TRADES \cite{zhang2019theoretically} has become one of the most favored methods, owing to its impressive experimental outcomes.

 While adversarially trained classifiers often exhibit higher overall robustness to input perturbations, it was discovered \cite{xu2021robust,li2021estimating,zhong2023learning} that they frequently create a significant disparity of standard accuracy and robust error between various classes or sensitive groups. For instance, a model that has been trained naturally exhibits comparable performance levels across different classes. Conversely, models that have undergone adversarial training tend to perform well in certain classes but perform worse in others. This disparity in performance poses significant risks for certain applications that rely on the model's accuracy and require comparable robustness across all classes. 
 To motivate this problem, consider the case of an autonomous driving system \cite{morgulis2019fooling,ma2022tradeoff} which employs an object detector. While this detector may exhibit an impressive level of robust accuracy when it comes to identifying objects on the road, there are certain classes that it struggles with. Specifically, it performs well when it comes to identifying inanimate objects, even with large perturbations, demonstrating a high level of robust accuracy on the ``inanimate" class. However, if it is less robust to perturbations while recognizing humans, this can be a significant safety concern. Researchers have observed a similar phenomenon in real-world datasets such as CIFAR-10 \cite{xu2021robust}. Specifically, models trained under normal conditions (i.e., minimizing the natural risk) tend to perform consistently across different classes. However, models trained using adversarial techniques exhibit significant discrepancies in performance by class, affecting both accuracy and robustness. For instance, such a model might demonstrate low errors—both standard and robust—on images from the ``car" class, while showing significantly higher error rates for ``cat" images. This issue of \textit{fairness}, characterized by uneven class-wise performance, is not as prominent in naturally trained models. 

  Recent studies by \cite{benz2021robustness,nanda2021fairness,xu2021robust,zhong2024intrinsic} have highlighted the issue of class-wise robustness disparity, which they characterize as a \textit{fair robustness} problem, emphasizing the need for equal robustness across different subgroups. \cite{xu2021robust} specifically notes that adversarial training (AT) can exacerbate disparities in standard accuracy across classes compared to natural training, and hence focuses on the development of adversarial training algorithms. In contrast, \cite{ma2022tradeoff} investigates how the perturbation radius in AT affects class-wise disparity in robust accuracy and examines whether there is a tradeoff between overall robust accuracy and class-wise disparity in robust accuracy. \cite{lee2024dafa} assigns suitable learning weights and adversarial margins to each class, based on the distances between classes.

  \textbf{Main Contributions} 
  In this paper, we investigate the fair robustness problem and propose the use of combining mixup with adversarial training as an effective solution to diminish class-wise risk discrepancies. We introduce a novel approach where we employ same-domain mixup samples—strategically using linear convex combinations of samples from same classes—to address disparities in natural risk. Similarly, for adversarial risk, we adapt this method by applying same domain mixup techniques with adversarial samples to equitably reduce risk across all classes. We present a detailed theoretical analysis for linear models and Gaussian data, which shows that using mixup \textit{provably} reduces class-wise disparities. Extensive empirical validation on both the synthetic dataset and the real-world dataset, such as CIFAR-10, substantiates our theoretical insights. For instance, on the CIFAR-10 dataset, the worst class test adversarial risk is about 12.70\% while the domain mixup worst class test adversarial risk is improved to 3.80\% when the perturbation budget $\epsilon = 0.3$. Our experiments confirm that our proposed method not only consistently reduces the discrepancies in class-wise risks but also significantly enhances the overall fairness and robustness of machine learning models. 

\section{Preliminaries and Problem Statement}
We consider a supervised classification problem, where we are given a dataset: $\{x_i, y_i\}_{i=1}^{n}$. We use $x_i \in \mathcal{X}$ to denote the set of features for $i$th sample and $y_i \in \mathcal{Y}$ denotes the corresponding true label, where $ \mathcal{X} \subseteq \mathbb{R}^{d}$ and $\mathcal{Y} = \{1,2,\dots, C\}$. Consider a classifier $f$ parameterized by $w$, which maps from input data space to output labels: $\mathcal{X} \rightarrow \mathcal{Y}$. We use $\ell(\cdot)$ to denote the loss function, e.g., $\ell(f_w(x),y)$. 

\vspace{5pt}
\noindent\textbf{Natural Training Classifier} The main objective is to optimize the model's accuracy on samples not previously seen from the base distribution. We define the natural training risk as follows
\begin{align}
    R_{\text{nat}}(f) = \mathbb{E}_{(x,y)}~ [ \ell(f_w(x),y)].
\end{align}
In addition, we identify the optimized natural training classifier $f_{\text{nat}}^*$ that has the lowest natural risk:
\begin{align}
    f_{\text{nat}}^* = \underset{f}{\text{argmin}}~R_{\text{nat}}(f).
\end{align}

\vspace{5pt}
\noindent\textbf{Adversarial Training Classifier} The goal of adversarial training is to construct models resistant to adversarial inputs through robust optimization techniques. In essence, adversarial training seeks to identify the classifier that exhibits the minimal adversarial or robust risk:
\begin{align}
    R_{\text{adv}}(f) = \mathbb{E}_{(x,y)}\left[ \underset{\parallel x' - x \parallel_\infty \leq \epsilon}{\text{sup}}\ell(f(x'),y) \right ].
\end{align}
where $\epsilon$ denotes the $\ell_\infty$ perturbation size. Similar as above, we define the optimized adversarial training classifier $f_{\text{adv}}^*$ that has the lowest adversarial risk:
\begin{align}
    f_{\text{adv}}^* = \underset{f}{\text{argmin}}~R_{\text{adv}}(f).
\end{align}

For the scope of this paper, we explore the class-wise discrepancies of both the natural training classifier and adversarial training classifier. Following previous works \cite{xu2021robust,ma2022tradeoff} and to facilitate the theoretical analysis, we consider the binary classification, $\mathcal{Y} = \{-1,1\}$, and the class-conditioned data are assumed to be drawn from a multi-variate Gaussian distribution as defined next.
 
\begin{definition}\label{def: mixgaussian} (Data Distribution)
Given some constants $\mu_{+}, \mu_{-}$ and $\sigma_{+}, \sigma_{-}$, the joint distribution of the data and labels are given as:
\begin{align}
y \hspace{-1pt}\sim \hspace{-1pt} \begin{cases}
    +1, & \text{w. p.}~ \alpha\\
    -1, & \text{w. p.}~ 1-\alpha\\
\end{cases},
~x \sim \begin{cases}
    \mathcal{N}(\bm{\mu_{+}}, \sigma_{+}^2I), & \text{if}~y = +1 \\
    \mathcal{N}(-\bm{\mu_{-}}, \sigma_{-}^2I), & \text{if}~y = -1 \\
\end{cases}\nonumber
\end{align}
where $\alpha$ is the prior probability of class +1 and $\bm{\mu_{+}}$ = $\mu_{+}\textbf{1}$, $\bm{\mu_{-}}$ = $\mu_{-}\textbf{1}$, $\textbf{1}$ represents the d-dimensional all-ones vector, and I is $d$-dimension identity matrix.
\end{definition}
Given a classifier $f$, we next define the class-wise natural risk with respect to the $0/1$ loss as follows:
\begin{align}
    &R_{\text{nat}}^{+1}(f) = \mathbb{E}_{(x,y)}(\mathbbm{1}(f(x)=-1|y=+1))\nonumber\\
    &R_{\text{nat}}^{-1}(f) = \mathbb{E}_{(x,y)}(\mathbbm{1}(f(x)=+1|y=-1))\nonumber.
\end{align}
where $\mathbbm{1}(\cdot)$ is the indicator function. To study the adversarial (robust) risk, we focus on $l_{\infty}$-norm bounded perturbations. Specifically, for an input $x$, the perturbation set around $x$ is defined as $\mathbb{B}(x, \epsilon)=\{x': \parallel x'-x\parallel_{\infty}\leq\epsilon\}$. We denote $\epsilon$ as the $l_{\infty}$ perturbation radius around the input. The class wise adversarial (or robust) risk can then be defined as:
\begin{align}
    &R_{\text{adv}}^{+1}(f) = \mathbb{E}_{(x,y)}\left(\underset{x'\in\mathbb{B}(x,\epsilon)}{\text{sup}}\mathbbm{1}(f(x')=-1|y=+1)\right)\nonumber\\
    &R_{\text{adv}}^{-1}(f) = \mathbb{E}_{(x,y)}\left(\underset{x'\in\mathbb{B}(x,\epsilon)}{\text{sup}}\mathbbm{1}(f(x')=+1|y=-1)\right).\nonumber
\end{align}
For the theoretical analysis, we assume that the classifier is linear, i.e.,  $f(x)= \text{sign}(\langle w,x\rangle+b)$, where $w \in \mathbb{R}^d$, $b \in \mathbb{R}$. Recent works, such as \cite{xu2021robust} and \cite{ma2022tradeoff}, have analyzed the class-wise natural and adversarial risks when $f$ is a linear model, identifying significant discrepancies among the classes. These results highlight the uneven distribution of natural risks (Proposition \ref{the:naturallinear}) and adversarial risks (Proposition \ref{the:adversa}) across different classes. 
\begin{proposition}\label{the:naturallinear} \cite{xu2021robust,ma2022tradeoff} (Class Wise Natural Risk)
For the optimized linear classifier $f^*_{\text{nat}}$ that minimizes the natural risk, when $\sigma_{-} = \sigma_{+} = \sigma$, the class wise risks are given as:
\begin{align}
    & R_{\text{nat}}^{+1}(f^*_{\text{nat}}) = \Phi \left( \frac{-d^2(\mu_{-}+\mu_{+})^2- 2K\sigma^2}{2d^{3/2}\sigma(\mu_{-}+\mu_{+})}\right), \\
    & R_{\text{nat}}^{-1}(f^*_{\text{nat}}) = \Phi \left( \frac{-d^2(\mu_{-}+\mu_{+})^2+ 2K\sigma^2}{2d^{3/2}\sigma(\mu_{-}+\mu_{+})}\right).
\end{align}
where K is a positive constant. 
\end{proposition}
Proposition \ref{the:naturallinear} shows that if the variance of two classes are the same, the risk of class +1 is less than the risk of class -1: $R_{\text{nat}}^{+1}(f^*_{\text{nat}}) < R_{\text{nat}}^{-1}(f^*_{\text{nat}})$, where the linear classifier shows bias since it prefers the positive class.
A similar result was obtained which showed the discrepancy of the class-wise robust risk. 
\begin{proposition}\label{the:adversa} \cite{xu2021robust,ma2022tradeoff} (Class Wise Adversarial Risk)
    For the optimized linear classifier $f^*_{\text{adv}}$, when $\sigma_{-} = \sigma_{+} = \sigma$, the class wise risks are given as:
\begin{align}
    & R_{\text{adv}}^{+1}(f^*_{\text{adv}}) = \Phi \left( \frac{-M'- 2K\sigma^2}{2d^{3/2}\sigma(\mu_{-}+\mu_{+}-2\epsilon)}\right), \\
    & R_{\text{adv}}^{-1}(f^*_{\text{adv}}) = \Phi \left( \frac{-M'+ 2K\sigma^2}{2d^{3/2}\sigma(\mu_{-}+\mu_{+}-2\epsilon)}\right).
\end{align}
where $M' = d^2(\mu_{-}+\mu_{+}-2\epsilon)\times(\mu_{-}+\mu_{+}-2\epsilon)$ and K is a positive constant.
\end{proposition}
Proposition \ref{the:adversa} shows that if the variance of two classes are the same, we have $R_{\text{adv}}^{+1}(f^*_{\text{adv}})<R_{\text{adv}}^{-1}(f^*_{\text{adv}})$, where the linear classifier shows bias in the class wise adversarial risk since it prefers the positive class. 

\vspace{5pt}
\noindent\textbf{Quantifying Class-wise Discrepancy:} for a given classifier $f$, in order to measure the classifier's bias present in the class-wise natural risks and class-wise adversarial risks, we define the difference between these risks as follows:
\begin{align}
    &\Delta_{\text{nat}}(f) \triangleq \left | R^{+}_{\text{nat}}(f) - R^{-}_{\text{nat}}(f) \right|, \nonumber\\
     &\Delta_{\text{adv}}(f) \triangleq \left | R^{+}_{\text{adv}}(f) - R^{-}_{\text{adv}}(f) \right|. \nonumber
\end{align}

\vspace{5pt}
\noindent\textbf{Problem Statement} Our goal is to build a robust classifier which diminishes the discrepancy of the class wise natural risks and adversarial risks. We propose to use the domain mixup mechanism to provably reduce the discrepancies of class wise natural risks as well as the adversarial risks. 
In the next section, we provide our main results and propose that mixup mechanisms can mitigate the class wise bias on either natural training or the adversarial training, for which we offer an in-depth theoretical analysis. 

\section{Main Results and Discussion}
In this section, we present our main results on mitigating the  class-wise discrepancies in natural risks and adversarial risks within the class-conditioned Gaussian distributions using a linear model. Our results are organized as follows: We first illustrate the mixup mechanisms that can reduce the gap among the class wise risk. We then give the theoretical analysis of the mixup based class wise natural risks (Theorem \ref{the:mixupnat}) and mixup based class wise adversarial risks (Theorem \ref{the: advmixp}) respectively. We establish that our mixup-based approach consistently reduces the discrepancies in class-wise risks, thereby leading to  a close to uniformly robust classifier.

\begin{figure*}[t]
    \centering
    \includegraphics[scale = 0.6]{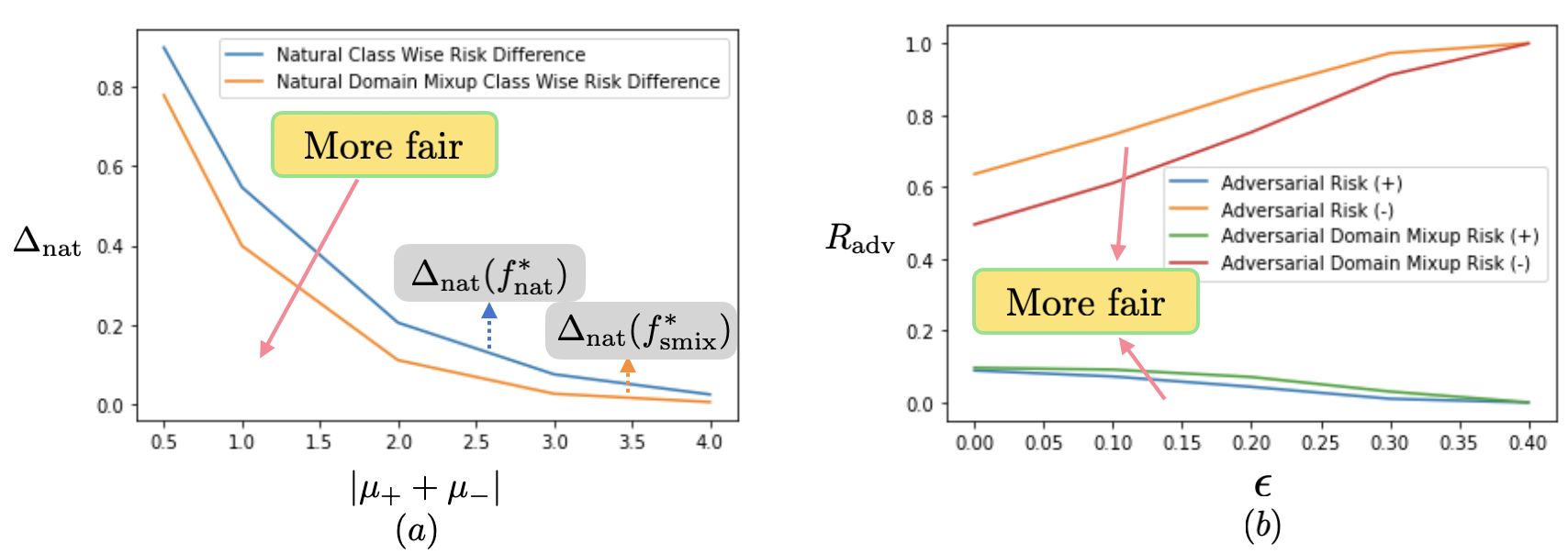}
    \caption{(a) Comparison of the natural risk and the natural domain mixup risk with the Gaussian data using a linear model with respect to the distance of two classes ($|\mu_{+}+\mu_{-}|$). We can observe that domain mixup mechanism consistently decreases the discrepancy of the class wise risk as we increase the distance (difference in mean) across the two classes. (b) Comparison of the adversarial risk and the adversarial domain mixup risk with the Gaussian data using a linear model as we increase the value of $\epsilon$. We can observe that domain mixup mechanism consistently reduces the gap of the class wise adversarial risk. }
    \label{fig: gaussian_comp}
\end{figure*}

The idea of Mixup was introduced in \cite{zhang2018mixup} as a data-dependent regularization technique that operates by taking linear combinations of pairs of examples and their labels to generate new training examples. This approach aims to encourage the model to behave more linearly in-between training examples, which can improve generalization and robustness of the model. Mixup has been shown to be effective in reducing generalization error across various tasks and models \cite{zhang2020does,verma2019manifold}. In our paper, we propose that mixup can decrease the discrepancy of the class wise risks and improve the fairness on the class wise natural risks and class wise adversarial risks.
We use $ x_{\text{smix}}$ denote the same domain mixup sample and $ y_{\text{smix}}$ denotes the same domain mixup label as defined below:
\begin{align}
    x_{\text{smix}} = \lambda x_i + (1-\lambda) x_j,~~y_{\text{smix}} =\lambda  y_i + (1-\lambda) y_j,
\end{align}
where $x_i, x_j$ are drawn from the same class and $i \neq j$.
Thus,  the joint distribution of the mixed up data $x_{\text{smix}}$, $y_{\text{smix}}$ (when the original data are Gaussian) can be written as follows:
\begin{align}
    &y_{\text{smix}} = +1~~ x_{\text{smix}}\sim \mathcal{N}(\bm{\mu_{+}}, (\lambda^2+ (1-\lambda)^2 )\sigma^2_{+}I),\\
    &y_{\text{smix}} = -1~~ x_{\text{smix}} \sim \mathcal{N}(-\bm{\mu_{-}}, (\lambda^2+ (1-\lambda)^2 )\sigma^2_{-}I).
\end{align}
where $\lambda \in [0,1]$. Considering the resulting variance of the mixup data, we can observe that the variance is proportional to $g(\lambda)= (\lambda^2+ (1-\lambda)^2 )$. Noting that $g(\lambda)$ is convex and $g(\lambda)$ is strictly less than $1$ for any $0<\lambda<1$, this shows that mixing up samples from the same distribution leads to a new distribution with reduced variance. 

\vspace{5pt}
\noindent\textbf{Domain Mixup Natural Training Classifier} The primary goal is to enhance the model's accuracy on new domain mixup samples that are derived from the base distribution but have not been previously encountered. Thus, we define the domain mixup natural risk as follows:
\begin{align}
    R_{\text{nat}}^{\text{smix}}(f) = \alpha \mathbb{E}_{(x_{\text{smix}},y_{\text{smix}})}~ [ \mathbbm{1}(f(x_{\text{smix}})=-1|y_{\text{smix}}=+1)\nonumber \\
    + (1-\alpha)\mathbb{E}_{(x_{\text{smix}},y_{\text{smix}})}~ [ \mathbbm{1}(f(x_{\text{smix}})=1|y_{\text{smix}}=-1) \nonumber
\end{align}
where $\alpha = \mathbb{P}(y_{\text{smix}} =1)$. 
We define the optimized same domain mixup classifier $f^*_{\text{smix}}$ that has the lowest domain mixup natural risk as:
\begin{align}
    f_{\text{smix}}^* = \underset{f}{\text{argmin}}~R_{\text{nat}}^{\text{smix}}(f)
\end{align}

\vspace{5pt}
 \noindent\textbf{Domain Mixup Adversarial Training Classifier} The aim of adversarial training is to build models that are resistant to adversarial inputs from the same domain mixup, using robust optimization methods. Therefore, the domain mixup adversarial risk is given as $R_{\text{adv}}^{\text{smix}} = $
 \begin{align}
    & \alpha \mathbb{E}\left[ \underset{\parallel x'-x_{\text{smix}}   \parallel_\infty \leq \epsilon}{\text{sup}} \mathbbm{1}(f(x')=-1|y_{\text{smix}}=+1)\right] +\nonumber \\
     &(1-\alpha)\mathbb{E} \left[\underset{\parallel x'-x_{\text{smix}}   \parallel_\infty \leq \epsilon}{\text{sup}} \mathbbm{1}(f(x')=1|y_{\text{smix}}=-1) \right].\nonumber
 \end{align}
 Similarly, the optimized adversarial same domain mixup classifier is:
 \begin{align}
     f^*_{\text{smix}} = \underset{f}{\text{argmin}}~R_{\text{adv}}^{\text{smix}}(f).
 \end{align}
We next present our first main result on domain mixup natural training.
\begin{theorem}\label{the:mixupnat} (Domain Mixup Natural Training)
    For the same domain mixup training linear classifier $f^*_{\text{smix}}$ that minimizes the natural risk, if $\sigma_{-} \neq \sigma_{+}$, the class wise risk is:
    \begin{align}
    & R_{\text{nat}}^{+1}(f^*_{\text{smix}}) = \Phi \left( \frac{-\eta^{*} - d\mu_{+}}{\sqrt{d(\lambda^2+ (1-\lambda)^2 )}\sigma_{+}}\right), \\
    & R_{\text{nat}}^{-1}(f^*_{\text{smix}}) = \Phi \left( \frac{\eta^{*} - d\mu_{-}}{\sqrt{d(\lambda^2+ (1-\lambda)^2 )}\sigma_{-}}\right)
\end{align}
where $\Phi(\cdot)$ is the cumulative distribution function (c.d.f) of normal distribution $\mathcal{N}$(0,1) and $\eta^{*} =$
\begin{align}
 \frac{-d(\mu_{+}\sigma^2_{-}+\mu_{-}\sigma^2_{+})+\sigma_{+} \sigma_{-}(\mu_{+}+\mu_{-})\sqrt{1+2K\frac{g(\lambda)(\sigma^2_{-}-\sigma^2_{+})}{(\mu_{-}+\mu_{+})^2}}}{\sigma^2_{-}-\sigma^2_{+}} \nonumber
\end{align}
where K is a positive constant and $g(\lambda) = \lambda^2+ (1-\lambda)^2 $. When $\sigma_{-} = \sigma_{+} = \sigma$, the class wise risk can be simplified as:
\begin{align}
    & R_{\text{nat}}^{+1}(f^*_{\text{smix}}) = \Phi \left( \frac{-d^2(\mu_{-}+\mu_{+})^2- 2K\sigma^2g(\lambda)}{2\sigma(\mu_{-}+\mu_{+})\sqrt{d^{3}g(\lambda)}}\right), \nonumber\\
    & R_{\text{nat}}^{-1}(f^*_{\text{smix}}) = \Phi \left( \frac{-d^2(\mu_{-}+\mu_{+})^2+ 2K\sigma^2g(\lambda)}{2\sigma(\mu_{-}+\mu_{+})\sqrt{d^{3}g(\lambda)}}\right). \nonumber
\end{align}
\end{theorem}
The proof of Theorem \ref{the:mixupnat} is presented in Appendix \ref{sec: theorem3}. We next present a sequence of remarks that give some operational interpretation behind the results and we also compare the class-wise discrepancy of mixup based adversarial training versus adversarial training (without mixup). 
\begin{remark}
   Theorem \ref{the:mixupnat} demonstrates that when the variance of two classes is identical, domain mixup training results in a reduced gap of risks between the classes . Specifically for the natural training, the difference of class risk is:
   \begin{align}
       &\Delta_{\text{nat}}(f^*_{\text{nat}}) = |R_{\text{nat}}^{+1}(f^*_{\text{nat}}) - R_{\text{nat}}^{-1}(f^*_{\text{nat}})| \nonumber\\
       &= \left | \Phi \left( \frac{- 2K\sigma^2}{2d^{3/2}\sigma(\mu_{-}+\mu_{+})}\right)-\Phi \left( \frac{ 2K\sigma^2}{2d^{3/2}\sigma(\mu_{-}+\mu_{+})}\right) \right | \nonumber \\
       & =\left | 2\Phi \left( \frac{- 2K\sigma^2}{2d^{3/2}\sigma(\mu_{-}+\mu_{+})}\right)  -1 \right |. \label{eq:natural_gap}
   \end{align}
   Similarly, for the domain mixup natural training, the difference of class risk is $ $
   \begin{align}
   &\Delta_{\text{nat}}(f^*_{\text{smix}}) = |R_{\text{nat}}^{+1}(f^*_{\text{smix}}) - R_{\text{nat}}^{-1}(f^*_{\text{smix}})| \nonumber\\
       &= \left | 2\Phi \left( \frac{- 2K\sigma^2\sqrt{g(\lambda)}}{2d^{3/2}\sigma(\mu_{-}+\mu_{+})}\right)-1\right |. \label{eq:natural_mixup_gap}
   \end{align}
   Comparing \eqref{eq:natural_gap} and \eqref{eq:natural_mixup_gap}, we have that:
   \begin{align}
       \Delta_{\text{nat}}(f^*_{\text{smix}}) \leq \Delta_{\text{nat}}(f^*_{\text{nat}}). \nonumber
   \end{align}
   Therefore, this result shows that domain mixup training decreases the discrepancy of the natural training class wise risk. We validate our results on Gaussian data as shown in Fig \ref{fig: gaussian_comp}(a).
\end{remark}
\begin{remark}
    Theorem \ref{the:mixupnat} states that if the variance of two classes are the same and $A \leq \lambda^2 + (1-\lambda)^2\leq 1$, we have:
\begin{align}
    R_{\text{nat}}^{+1}(f^*_{\text{nat}}) \leq R_{\text{nat}}^{+1}(f^*_{\text{smix}}) \leq R_{\text{nat}}^{-1}(f^*_{\text{smix}})
     \leq R_{\text{nat}}^{-1}(f^*_{\text{nat}}). \nonumber
\end{align}
where $A = d^2(\mu_{+}+\mu_{-})^4 /4K^2\sigma^4 $. From above, the domain mixup training can balance the class wise risk and further validate the effectiveness of Remark 1. 
\end{remark}
\begin{figure*}[t]
    \centering
    \includegraphics[scale = 0.6]{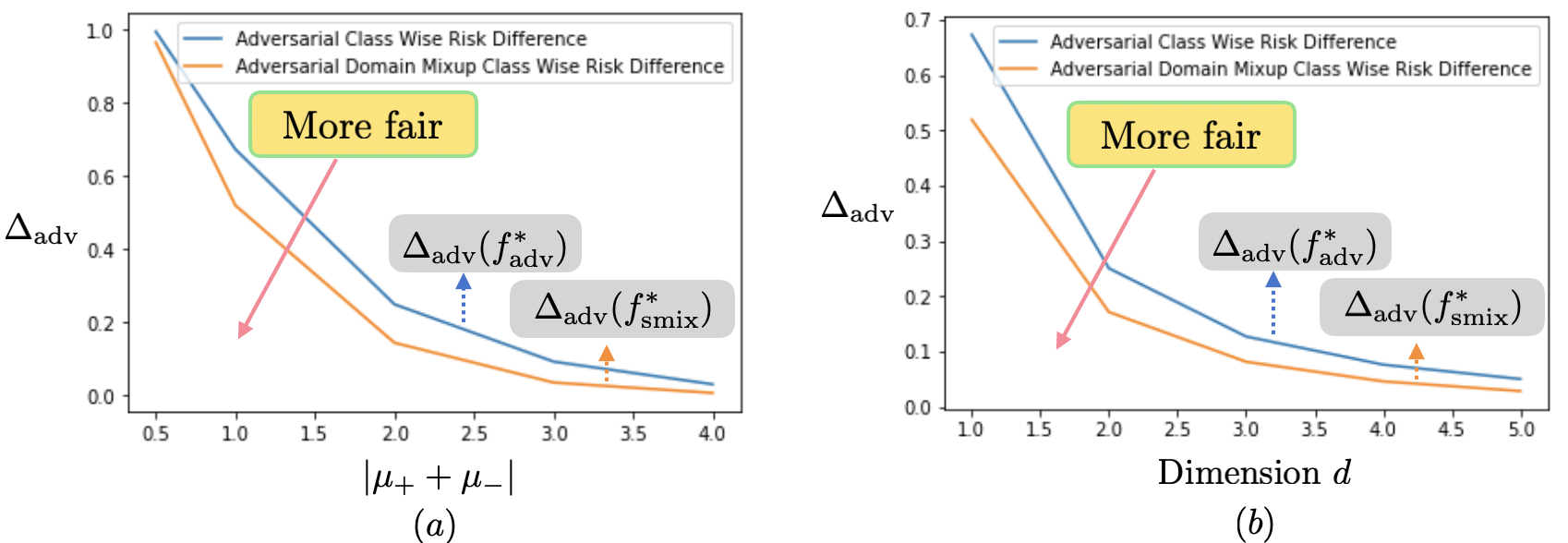}
    \caption{(a) Comparison of the adversarial risk and the adversarial domain mixup risk with the Gaussian data using a linear model with respect to the distance of two classes ($|\mu_{+}+\mu_{-}|$). We can observe that domain mixup mechanism consistently decreases the discrepancy of the class wise risk as we increase the distance of two classes. (b) Comparison of the adversarial risk and the adversarial domain mixup risk with the Gaussian data using a linear model as we increase the dimension d. We note that Mixup mechanism consistently reduces the gap of the class wise adversarial risk when we increase the data dimension. }
    \label{fig: gaussian_comp_2}
\end{figure*}

We then analyze the same domain mixup adversarial training, where we mixup the same class adversarial samples. We then present our findings in the remarks that domain mixup training process consistently mitigates the class wise adversarial risk gap. 
\begin{theorem}\label{the: advmixp}
    For the same domain mixup adversarial training linear classifier $f^*_{\text{smix}}$, if $\sigma_{-}\neq\sigma_{+}$, its corresponding class wise adversarial risk is:
\begin{align}
    & R_{\text{adv}}^{+1}(f^*_{\text{smix}}) = \Phi \left( \frac{-s^{*} - d(\mu_{+}-\epsilon)}{\sqrt{d g(\lambda)}\sigma_{+}}\right), \nonumber \\
    & R_{\text{adv}}^{-1}(f^*_{\text{smix}}) = \Phi \left( \frac{s^{*} - d(\mu_{-}-\epsilon)}{\sqrt{d g(\lambda)}\sigma_{-}}\right)
\end{align}
where $\Phi(\cdot)$ is the cumulative distribution function (c.d.f) of normal distribution $\mathcal{N}$(0,1) and $s^{*} =$
\begin{align}
 \frac{M+\sigma_{+} \sigma_{-}(\mu_{+}+\mu_{-}-2\epsilon)\sqrt{1+2Kg(\lambda)\frac{\sigma^2_{-}-\sigma^2_{+}}{(\mu_{-}+\mu_{+}-2\epsilon)^2}}}{\sigma^2_{-}-\sigma^2_{+}}
\end{align}
where $M=-d(\mu_{+}\sigma^2_{-}+\mu_{-}\sigma^2_{+}-\epsilon(\sigma_{+}^2+\sigma_{-}^2))$ and K is a positive constant. When $\sigma_{-} = \sigma_{+} = \sigma$, the class wise risk is:
\begin{align}
    R_{\text{adv}}^{+1}(f^*_{\text{smix}}) = \Phi \left( \frac{-M'- 2K\sigma^2g(\lambda)}{2\sigma(\mu_{-}+\mu_{+}-2\epsilon)\sqrt{d^{3} g(\lambda)}}\right)\\
    R_{\text{adv}}^{-1}(f^*_{\text{smix}}) = \Phi \left( \frac{-M'+ 2K\sigma^2g(\lambda)}{2\sigma(\mu_{-}+\mu_{+}-2\epsilon)\sqrt{d^{3} g(\lambda)}}\right)
\end{align}
where $M' = d^2(\mu_{-}+\mu_{+}-2\epsilon)\times(\mu_{-}+\mu_{+}-2\epsilon)$. 
\end{theorem}
We provide the proof in Appendix \ref{sec: theorem4}. 
\begin{remark}
    Theorem \ref{the: advmixp} demonstrates that when the variance of two classes are the same, domain mixup adversarial training can decrease the discrepancy between the classes. For the adversarial risk, the difference of the adversarial class risk is
    \begin{align}
        \Delta_{\text{adv}}(f^*_{\text{adv}}) = \left | 2\Phi \left( \frac{- 2K\sigma^2}{2d^{3/2}\sigma(\mu_{-}+\mu_{+}-2\epsilon}\right)  -1 \right |. \label{eq: adver_risk_gap}
    \end{align}
    Similarly, for the domain mixup adversarial risk, the the difference of the class risk is
    \begin{align}
        \Delta_{\text{adv}}(f^*_{\text{smix}}) = \left | 2\Phi \left( \frac{- 2K\sigma^2 \sqrt{g(\lambda)}}{2d^{3/2}\sigma(\mu_{-}+\mu_{+}-2\epsilon}\right)  -1 \right |. \label{eq: adver_risk_mix_gap}
    \end{align}
    Combining \eqref{eq: adver_risk_gap} and \eqref{eq: adver_risk_mix_gap}, we arrive at:
    \begin{align}
        \Delta_{\text{adv}}(f^*_{\text{smix}})  \leq \Delta_{\text{adv}}(f^*_{\text{adv}}). \nonumber
    \end{align}
    We validate our results on Gaussian data as shown in Fig \ref{fig: gaussian_comp}(b) and \ref{fig: gaussian_comp_2}. We show that domain mixup can consistently mitigate the discrepancy of the adversarial class wise risks.
\end{remark}
\begin{remark}
    Theorem \ref{the: advmixp} shows that if the variance of two classes are the same and $B\leq g(\lambda) \leq 1$, we have:
    \begin{align}
        R_{\text{adv}}^{+1}(f^*_{\text{adv}})  \leq R_{\text{adv}}^{+1}(f^*_{\text{smix}}) \leq  R_{\text{adv}}^{-1}(f^*_{\text{smix}}) \leq R_{\text{adv}}^{-1}(f^*_{\text{adv}}). \nonumber
    \end{align}
    where $B = M'^2 /4K^2\sigma^4$. The above result demonstrates that the same domain mixup training can balance the adversarial risk on each class, which can reduce the bias in the adversarial training classifier.
\end{remark}

\begin{table*}[t]
    \centering
    \begin{tabular}{| l | c |c| c| c| c|}
    \hline
      & Class Risk Avg  &  Class Risk Std  & Class Risk Min & Class Risk Max & $\Delta_{adv}$ \\
      \hline
      Adversarial Risk ($\epsilon = 0.1$)  & 0.0471 & 0.0298 &  0.0130 & 0.1240 & 0.1110\\
       \hline
       Domain Mixup Adversarial Risk ($\epsilon = 0.1$)  & \textbf{0.0484} & \textbf{0.0239} &\textbf{0.0190 }& \textbf{0.0890 } & \textbf{0.0700} \\
       \hline
       Adversarial Risk ($\epsilon = 0.3$)  & 0.0698 & 0.0337 &  0.0170 & 0.1270 & 0.1100\\
       \hline
       Domain Mixup Adversarial Risk ($\epsilon = 0.3$)  & \textbf{0.0113} & \textbf{0.0131} & \textbf{0.0000} & \textbf{0.0380} & \textbf{0.0380}\\
       \hline
    \end{tabular}
    \caption{Comparison of the adversarial risk and the domain mixup adversarial risk on the CIFAR-10 test dataset, where ``avg" denotes the mean, ``std" represents the standard deviation, and ``min" (``max") indicates the minimum (maximum) value among the class-wise risks.}
    \label{tab:class_wise_stat}
\end{table*}

\noindent \textbf{Evaluation of Mixup AT on CIFAR-10} In this section, we validate the proposed approach of mixup on CIFAR-10 dataset using a more complex model (ResNet20). 

\noindent \textit{Dataset and Classifier}
The CIFAR-10 dataset comprises 60,000 color images distributed across 10 classes: airplane, automobile, bird, cat, deer, dog, frog, horse, ship, and truck, with each class containing 6,000 images. The dataset is divided into a training set of 50,000 images and a test set of 10,000 images. Images are in RGB format, necessitating the construction of three input channels for the CIFAR-10 dataset. Accuracy metrics are reported on the test dataset. To enhance performance, we preprocess the CIFAR-10 dataset using normalization; each image's channel mean (0.4914, 0.4822, 0.4465) is subtracted and then divided by its standard deviation (0.2023, 0.1994, 0.2010).

We employ a Residual Network (ResNet 20) to achieve high accuracy, where "20" denotes the number of layers in the architecture (other variants include 18, 34, etc.). ResNet, a CNN-based architecture, is designed to facilitate much deeper neural networks. Traditional deep CNN models often encounter the problem of vanishing gradients, where repeated multiplication of weights leads to the gradual disappearance of gradients. ResNet addresses this by implementing a technique known as "skip connection", which involves stacking identity layers, skipping them, and then reusing these layers when additional feature spaces are required.

We apply the Fast Gradient Sign Method (FGSM) \cite{goodfellow2014explaining} as the adversarial attack technique, which aims to generate adversarial examples by perturbing legitimate inputs using the gradient information of the target model. Specifically, it computes the gradient of the model's loss function with respect to the input and determine the sign of the gradient, indicating the direction that maximally increases the loss. Then it multiplies the sign by a small epsilon value to determine the perturbation magnitude and add the perturbation to the original input to create the adversarial example. During the training process, the batch size is $256$. For the learning rate, we initiate at $0.001$ and subsequently reduce it by a factor of $0.1$ every 50 epochs. The total training epoch is set at 80 epochs. We use Adam optimizer during the training process.

As shown in Table \ref{tab:class_wise_stat}, we can observe that the worst class test adversarial risk is about 12.70\% while the domain mixup worst class test adversarial risk is improved to 3.80\% when $\epsilon = 0.3$. We can also observe that the standard variation of the class wise domain mixup adversarial risk consistently outperform the adversarial training  risk. For instance, when $\epsilon = 0.3$, the standard deviation of the class wise risk on domain mixup adversarial training is $0.1310$, where standard deviation of the class wise risk on adversarial training is $0.0337$.
\section{Conclusion}

In conclusion, this paper addresses the fair robustness problem in adversarial training (AT) by proposing the same domain mixup mechanism in developing fair and robust classifiers. We specifically propose mixup inputs from same classes and mixup the adversarial samples from same classes to reduce the discrepancy in class wise natural risks and adversarial risks respectively. Furthermore, we provide a theoretical analysis for domain mixup mechanisms, demonstrating that combining mixup with adversarial training can reduce class-wise robustness disparities. This approach not only fosters less discrepancies in adversarial risks but also promotes uniformity in the results of natural training processes. Additionally, we present experimental evidence from both synthetic data and real-world datasets like CIFAR-10, which consistently shows improvements in reducing class-wise disparities for both natural and adversarial risks.

\bibliographystyle{ieeetr}
\bibliography{reference}

\begin{thebibliography}{10}

\bibitem{goodfellow2014explaining}
I.~J. Goodfellow, J.~Shlens, and C.~Szegedy, ``Explaining and harnessing adversarial examples,'' {\em arXiv preprint arXiv:1412.6572}, 2014.

\bibitem{szegedy2013intriguing}
C.~Szegedy, W.~Zaremba, I.~Sutskever, J.~Bruna, D.~Erhan, I.~Goodfellow, and R.~Fergus, ``Intriguing properties of neural networks,'' {\em arXiv preprint arXiv:1312.6199}, 2013.

\bibitem{morgulis2019fooling}
N.~Morgulis, A.~Kreines, S.~Mendelowitz, and Y.~Weisglass, ``Fooling a real car with adversarial traffic signs,'' {\em arXiv preprint arXiv:1907.00374}, 2019.

\bibitem{madry2017towards}
A.~Madry, A.~Makelov, L.~Schmidt, D.~Tsipras, and A.~Vladu, ``Towards deep learning models resistant to adversarial attacks,'' {\em arXiv preprint arXiv:1706.06083}, 2017.

\bibitem{pang2020boosting}
T.~Pang, X.~Yang, Y.~Dong, K.~Xu, J.~Zhu, and H.~Su, ``Boosting adversarial training with hypersphere embedding,'' {\em Advances in Neural Information Processing Systems}, vol.~33, pp.~7779--7792, 2020.

\bibitem{zhang2019theoretically}
H.~Zhang, Y.~Yu, J.~Jiao, E.~Xing, L.~El~Ghaoui, and M.~Jordan, ``Theoretically principled trade-off between robustness and accuracy,'' in {\em International conference on machine learning}, pp.~7472--7482, PMLR, 2019.

\bibitem{carmon2019unlabeled}
Y.~Carmon, A.~Raghunathan, L.~Schmidt, J.~C. Duchi, and P.~S. Liang, ``Unlabeled data improves adversarial robustness,'' {\em Advances in neural information processing systems}, vol.~32, 2019.

\bibitem{schmidt2018adversarially}
L.~Schmidt, S.~Santurkar, D.~Tsipras, K.~Talwar, and A.~Madry, ``Adversarially robust generalization requires more data,'' {\em Advances in neural information processing systems}, vol.~31, 2018.

\bibitem{wong2020fast}
E.~Wong, L.~Rice, and J.~Z. Kolter, ``Fast is better than free: Revisiting adversarial training,'' {\em arXiv preprint arXiv:2001.03994}, 2020.

\bibitem{dobriban2020provable}
E.~Dobriban, H.~Hassani, D.~Hong, and A.~Robey, ``Provable tradeoffs in adversarially robust classification,'' {\em arXiv preprint arXiv:2006.05161}, 2020.

\bibitem{javanmard2020precise}
A.~Javanmard, M.~Soltanolkotabi, and H.~Hassani, ``Precise tradeoffs in adversarial training for linear regression,'' in {\em Conference on Learning Theory}, pp.~2034--2078, PMLR, 2020.

\bibitem{zhong2024splitz}
M.~Zhong and R.~Tandon, ``Splitz: Certifiable robustness via split lipschitz randomized smoothing,'' {\em arXiv preprint arXiv:2407.02811}, 2024.

\bibitem{zhang2024filtered}
W.~Zhang, M.~Zhong, R.~Tandon, and M.~Krunz, ``Filtered randomized smoothing: A new defense for robust modulation classification,'' {\em arXiv preprint arXiv:2410.06339}, 2024.

\bibitem{xu2021robust}
H.~Xu, X.~Liu, Y.~Li, A.~Jain, and J.~Tang, ``To be robust or to be fair: Towards fairness in adversarial training,'' in {\em International Conference on Machine Learning}, pp.~11492--11501, PMLR, 2021.

\bibitem{li2021estimating}
X.~Li, Z.~Cui, Y.~Wu, L.~Gu, and T.~Harada, ``Estimating and improving fairness with adversarial learning,'' {\em arXiv preprint arXiv:2103.04243}, 2021.

\bibitem{zhong2023learning}
M.~Zhong and R.~Tandon, ``Learning fair classifiers via min-max f-divergence regularization,'' in {\em 2023 59th Annual Allerton Conference on Communication, Control, and Computing (Allerton)}, pp.~1--8, IEEE, 2023.

\bibitem{ma2022tradeoff}
X.~Ma, Z.~Wang, and W.~Liu, ``On the tradeoff between robustness and fairness,'' in {\em Advances in Neural Information Processing Systems}, 2022.

\bibitem{benz2021robustness}
P.~Benz, C.~Zhang, A.~Karjauv, and I.~S. Kweon, ``Robustness may be at odds with fairness: An empirical study on class-wise accuracy,'' in {\em NeurIPS 2020 Workshop on Pre-registration in Machine Learning}, pp.~325--342, PMLR, 2021.

\bibitem{nanda2021fairness}
V.~Nanda, S.~Dooley, S.~Singla, S.~Feizi, and J.~P. Dickerson, ``Fairness through robustness: Investigating robustness disparity in deep learning,'' in {\em Proceedings of the 2021 ACM Conference on Fairness, Accountability, and Transparency}, pp.~466--477, 2021.

\bibitem{zhong2024intrinsic}
M.~Zhong and R.~Tandon, ``Intrinsic fairness-accuracy tradeoffs under equalized odds,'' {\em arXiv preprint arXiv:2405.07393}, 2024.

\bibitem{lee2024dafa}
H.~Lee, S.~Lee, H.~Jang, J.~Park, H.~Bae, and S.~Yoon, ``Dafa: Distance-aware fair adversarial training,'' {\em arXiv preprint arXiv:2401.12532}, 2024.

\bibitem{zhang2018mixup}
H.~Zhang, M.~Cisse, Y.~N. Dauphin, and D.~Lopez-Paz, ``mixup: Beyond empirical risk minimization,'' in {\em International Conference on Learning Representations}, 2018.

\bibitem{zhang2020does}
L.~Zhang, Z.~Deng, K.~Kawaguchi, A.~Ghorbani, and J.~Zou, ``How does mixup help with robustness and generalization?,'' {\em arXiv preprint arXiv:2010.04819}, 2020.

\bibitem{verma2019manifold}
V.~Verma, A.~Lamb, C.~Beckham, A.~Najafi, I.~Mitliagkas, D.~Lopez-Paz, and Y.~Bengio, ``Manifold mixup: Better representations by interpolating hidden states,'' in {\em International conference on machine learning}, pp.~6438--6447, PMLR, 2019.

\end{thebibliography}
\appendices
\section{Proof of Theorem \ref{the:mixupnat}} \label{sec: theorem3}
\begin{proof}
    For any arbitrary linear classifier $f(x_{\text{smix}})= \text{sign}(\langle w, x_{\text{smix}} \rangle+b)$, its same domain mixup training risk $R_{\text{nat}}(f)$ is:
    \begin{align}
    &= \mathbb{E}_{(x_{\text{smix}},y_{\text{smix}})}(\mathbbm{1}(f(x_{\text{smix}})\neq y_{\text{smix}}))= \mathbb{P}_{(x,y)}(f(x_{\text{smix}})\neq y_{\text{smix}})\nonumber\\
    & = \mathbb{P}(y_{\text{smix}} =1) \times \mathbb{P}(f(x_{\text{smix}}) = -1 |y_{\text{smix}}=1) \nonumber\\
    &+ \mathbb{P}(y_{\text{smix}} =-1) \times \mathbb{P}(f(x_{\text{smix}}) = 1 | y_{\text{smix}}=-1).  
    \end{align}
    Let $\alpha = \mathbb{P}(y_{\text{smix}} =1)$, then the same domain mixup training risk becomes:
    \begin{align}
        R_{\text{nat}}(f) = \alpha \times R_{\text{nat}}^{+1}(f) + (1-\alpha) \times R_{\text{nat}}^{-1}(f). \nonumber
    \end{align}
    Note that $x_{\text{smix}}, w\in \mathbb{R}^d$ and $b \in \mathbb{R}$, we then expend the risk of class +1 as $R_{\text{nat}}^{+1}(f_{\text{smix}}) = $
    \begin{align}
        \mathbb{P}(f(x_{\text{smix}}) = -1 |y_{\text{smix}}=1) = \mathbb{P}(\langle w, x_{\text{smix}} \rangle + b <0 |y_{\text{smix}}=1). \nonumber
    \end{align}
    Since $x_{\text{smix}} = (x_{\text{smix}}^1, \dots, x_{\text{smix}}^d)$ are drawn from Gaussian distribution $\mathcal{N}(\bm{\mu_{+}}, (\lambda^2+ (1-\lambda)^2 )\sigma_{+}^2I)$, we have:
    \begin{align}
        R_{\text{nat}}^{+1} = \mathbb{P}(\sum_{i=1}^d w_i x_{\text{smix}}^i +b < 0).
    \end{align}
    Following the strategy of proof by contradiction (as in \cite{xu2021robust}) we can prove that the optimal weights must satisfy $w_1^* = w_2^* = \dots = w_d^* $. Let $w^* = w_i^*$, then we can simplify the risk as $R_{\text{nat}}^{+1}(f)=\mathbb{P}(w^*\sum_{i=1}^d x_{\text{smix}}^i +b < 0)$
    \begin{align}
        &  =\mathbb{P}\left (\frac{w^*\sum_{i=1}^d (x_{\text{smix}}^i-\mu_{+})}{\sqrt{\sum_{i=1}^{d}(w^*)^2\sigma_{+}^2g(\lambda)}}
         < \frac{-b-dw^*\mu_{+}}{\sqrt{\sum_{i=1}^{d}(w^*)^2\sigma_{+}^2g(\lambda)}}\right)\nonumber\\
        & = \Phi\left(\frac{-b-dw^*\mu_{+}}{\sqrt{dg(\lambda)} w^*\sigma_{+}}\right).
    \end{align}
    Following the same steps as above, we can derive the risk of class -1 as $R_{\text{nat}}^{-1}(f) == \mathbb{P}(w^*\sum_{i=1}^d x_{\text{smix}}^i +b > 0)$
    \begin{align}
        & = \mathbb{P}\left(\frac{w^*\sum_{i=1}^d (x_{\text{smix}}^i+\mu_{-})}{\sqrt{\sum_{i=1}^{d}(w^*)^2\sigma_{-}^2g(\lambda)}}>\frac{-b+dw^*\mu_{-}}{\sqrt{\sum_{i=1}^{d}(w^*)^2\sigma_{-}^2g(\lambda)}}\right)\nonumber\\
        & = \Phi\left(\frac{b-dw^*\mu_{-}}{\sqrt{dg(\lambda)}w^*\sigma_{-}}\right).
    \end{align}
    There the overall natural risk $R_{\text{nat}}(f_{\text{smix}}) = $
    \begin{align}
         \alpha\Phi\left(\frac{-b/w^*-d\mu_{+}}{\sqrt{dg(\lambda)}\sigma_{+}}\right) \nonumber+ (1-\alpha)\Phi\left(\frac{b/w^*-d\mu_{-}}{\sqrt{dg(\lambda)}\sigma_{-}}\right).\label{eq:overallnat_mix}
    \end{align}
    Let $t^* =b^*/w^*$ and take the derivative of equation \eqref{eq:overallnat_mix} w.r.t $t$. To find the optimal value of $t$, we set the derivative to zero:
    \begin{align}
        &\frac{\alpha}{\sqrt{2\pi}}\text{exp}\left(-\frac{1}{2}\left(\frac{-t^*-d\mu_{+}}{\sqrt{dg(\lambda)}\sigma_{+}}\right)^2\right)\times\frac{-1}{\sqrt{dg(\lambda)}\sigma_{+}} \nonumber \\
        &+ \frac{1-\alpha}{\sqrt{2\pi}}\text{exp}\left(-\frac{1}{2}\left(\frac{t^*-d\mu_{-}}{\sqrt{dg(\lambda)}\sigma_{-}}\right)^2\right)\times\frac{1}{\sqrt{dg(\lambda)}\sigma_{-}} = 0 \nonumber
    \end{align}
    Solving the above equation, we find that $t^*$ can be expressed as:
    \begin{align}
        &t^* = \frac{-d(\mu_{+}\sigma^2_{-}+\mu_{-}\sigma^2_{+})}{\sigma^2_{-}-\sigma^2_{+}} \nonumber\\
        &+ \frac{\sigma_{+} \sigma_{-}(\mu_{+}+\mu_{-})\sqrt{1+2Kg(\lambda)\frac{\sigma^2_{-}-\sigma^2_{+}}{(\mu_{-}+\mu_{+})^2}}}{\sigma^2_{-}-\sigma^2_{+}}
    \end{align}
    where $K= d\text{log}\left(\frac{\alpha \sigma_{-}}{(1-\alpha)\sigma_{+}}\right)$. When $\sigma_{+} = \sigma_{-} = \sigma$, we have:
    \begin{align}\label{eq:solutiont}
        t^* = \frac{-d^2(\mu_{+}^2-\mu_{-}^2)+2K\sigma^2g(\lambda)}{2d(\mu_{+}+\mu_{-})}.
    \end{align}
    We plug $t^*$ into equation \eqref{eq:overallnat_mix} and class wise risk becomes:
    \begin{align}
        & R_{\text{nat}}^{+1}(f^*_{\text{smix}}) = \Phi \left( \frac{-d^2(\mu_{-}+\mu_{+})^2- 2K\sigma^2g(\lambda)}{2\sigma(\mu_{-}+\mu_{+})\sqrt{d^{3}g(\lambda)}}\right), \nonumber \\
    & R_{\text{nat}}^{-1}(f^*_{\text{smix}}) = \Phi \left( \frac{-d^2(\mu_{-}+\mu_{+})^2+ 2K\sigma^2g(\lambda)}{2\sigma(\mu_{-}+\mu_{+})\sqrt{d^{3}g(\lambda)}}\right).
    \end{align}
    This completes the proof of Theorem \ref{the:mixupnat}. 
\end{proof}
\section{Proof of Theorem \ref{the: advmixp}}\label{sec: theorem4}
\begin{proof}
      For any arbitrary linear classifier $f= \text{sign}(\langle w,x_{\text{smix}}\rangle+b)$,
    the adversarial risk becomes:
    \begin{align}
        R_{\text{adv}}(f) = \alpha R_{\text{adv}}^{+1}(f) + (1-\alpha)  R_{\text{adv}}^{-1}(f) \nonumber
    \end{align}
    Note that $x, w\in \mathbb{R}^d$ and $b,\epsilon \in \mathbb{R}$.
    From the paper \cite{xu2021robust}, we can proof $w_1^* = w_2^* = \dots = w_d^* $ by contradiction. Let $w^* = w_i^*$, then we can simplify the risk as follows:
    \begin{align}
        R_{\text{adv}}^{+1} &= \mathbb{P}(w^*\sum_{i=1}^d (x_{\text{smix}}^i -\epsilon)+b < 0)\\
        & = \Phi\left(\frac{-b-dw^*(\mu_{+}-\epsilon)}{\sqrt{dg(\lambda)}w^*\sigma_{+}}\right). \label{eq: adv_mix_pos}
    \end{align}
    Following the same steps as above, we can derive the risk of class -1 as follows:
    \begin{align}
        R_{\text{adv}}^{-1} &= \mathbb{P}(w^*\sum_{i=1}^d (x_{\text{smix}}^i+\epsilon) +b > 0)\\
        & = \Phi\left(\frac{b-dw^*(\mu_{-}-\epsilon)}{\sqrt{dg(\lambda)}w^*\sigma_{-}}\right). \label{eq: adv_mix_neg}
    \end{align}
    There the overall natural risk $R_{\text{adv}}(f^*_{\text{smix}}) = $
    \begin{align}
         \alpha\Phi\left(\frac{\frac{-b}{w^*}-d(\mu_{+}-\epsilon)}{\sqrt{dg(\lambda)}\sigma_{+}}\right) + (1-\alpha)\Phi\left(\frac{\frac{b}{w^*}-d(\mu_{-}-\epsilon)}{\sqrt{dg(\lambda)}\sigma_{-}}\right).\label{eq:overalladv_mix}
    \end{align}
    Let $s^* =b^*/w^*$ and take the derivative of equation \eqref{eq:overalladv_mix} w.r.t $t$. To find the optimal value of $t$, we set the derivative to zero.
    We find that $s^*$ can be expressed as:
    \begin{align}
        \frac{M+\sigma_{+} \sigma_{-}(\mu_{+}+\mu_{-}-2\epsilon)\sqrt{1+2K\frac{(\sigma^2_{-}-\sigma^2_{+})\cdot g(\lambda)}{(\mu_{-}+\mu_{+}-2\epsilon)^2}}}{\sigma^2_{-}-\sigma^2_{+}} \nonumber
    \end{align}
    where $M=-d(\mu_{+}\sigma^2_{-}+\mu_{-}\sigma^2_{+}-\epsilon(\sigma_{+}^2+\sigma_{-}^2))$, $K= d\text{log}\left(\frac{\alpha \sigma_{-}}{(1-\alpha)\sigma_{+}}\right)$. When $\sigma_{+} = \sigma_{-} = \sigma$, we have:
    \begin{align}\label{eq:solutiont}
        s^* = \frac{-d^2((\mu_{+}-\epsilon)^2-(\mu_{-}-\epsilon)^2)+2K\sigma^2g(\lambda)}{2d(\mu_{+}+\mu_{-}-2\epsilon)}
    \end{align}
    where $K= d\text{log}\left(\frac{\alpha }{(1-\alpha)}\right)$. 
    We plug $s^*$ into \eqref{eq: adv_mix_pos} and \eqref{eq: adv_mix_neg}, then class wise risk is:
    \begin{align}
        R_{\text{adv}}^{+1}(f^*_{\text{smix}}) = \Phi \left( \frac{-M'- 2K\sigma^2g(\lambda)}{2\sigma(\mu_{-}+\mu_{+}-2\epsilon)\sqrt{d^3 g(\lambda)}}\right)\\
    R_{\text{adv}}^{-1}(f^*_{\text{smix}}) = \Phi \left( \frac{-M'+ 2K\sigma^2g(\lambda)}{2\sigma(\mu_{-}+\mu_{+}-2\epsilon)\sqrt{d^3 g(\lambda)}}\right),
    \end{align}
    where $M' = d^2(\mu_{-}+\mu_{+}-2\epsilon)\times(\mu_{-}+\mu_{+}-2\epsilon)$.
    This completes the proof of Theorem \ref{the: advmixp}. 
\end{proof}
\end{document}